# A Novel Representation of Periodic Pattern and Its Application to Untrained Anomaly Detection

Peng Ye [a], Chengyu Tao [b], Juan Du [abcd]*

[a] *Smart Manufacturing Thrust, The Hong Kong University of Science and Technology (Guangzhou), Guangzhou, China*
[b] *Division of Emerging Interdisciplinary Areas, The Hong Kong University of Science and Technology, Hong Kong SAR, China*
[c] *Department of Mechanical and Aerospace Engineering, The Hong Kong University of Science and Technology, Hong Kong SAR, China*
[d] *Guangzhou HKUST Fok Ying Tung Research Institute, Guangzhou, China*
*Corresponding Author, Juan Du, juandu@ust.hk

**Abstract**

There are a variety of industrial products that possess periodic textures or surfaces, such as carbon fiber textiles and display panels. Traditional image-based quality inspection methods for these products require identifying the periodic patterns from normal images (without anomaly and noise) and subsequently detecting anomaly pixels with inconsistent appearances. However, it remains challenging to accurately extract the periodic pattern from a single image in the presence of unknown anomalies and measurement noise. To deal with this challenge, this paper proposes a novel self-representation of the periodic image defined on a set of continuous parameters. In this way, periodic pattern learning can be embedded into a joint optimization framework, which is named periodic-sparse decomposition, with simultaneously modeling the sparse anomalies and Gaussian noise. Finally, for the real-world industrial images that may not strictly satisfy the periodic assumption, we propose a novel pixel-level anomaly scoring strategy to enhance the performance of anomaly detection. Both simulated and real-world case studies demonstrate the effectiveness of the proposed methodology for periodic pattern learning and anomaly detection.

# 1. Introduction

A variety of industrial products feature periodic textures and surfaces, including carbon fiber textiles (Szarski and Chauhan, 2022; Zambal et al., 2015; Zhang et al., 2018), periodic texture textiles (Ngan et al., 2008), and display panels (Çelik et al., 2022; Kim et al., 2020). In the product manufacturing process, operational and mechanical failures can lead to anomalies on the product surface. As a result, ensuring effective inspection and control of product quality on these surfaces is crucial within the modern manufacturing industry.

In the field of anomaly detection for product surfaces, recent research has tended to utilize training-based methods. However, the collection and labeling of training datasets for each specific texture can be both labor-intensive and costly due to the wide range of patterns and textures observed on product surfaces. As a result, untrained anomaly detection methods that analyze a single input sample and do not require additional training data have attracted significant attention.

Most existing untrained anomaly detection methods focus on simple product surfaces, such as smooth surfaces (Tao and Du, 2023; Tao et al., 2023; Yan et al., 2017). However, these methods cannot be applied to non-smooth surfaces commonly in manufacturing products, such as general axis-symmetric surfaces and periodic surfaces. For axis-symmetric surfaces, Cao et al. proposed an untrained anomaly detection method based on robust principal component analysis (Cao et al., 2024). Several studies have also proposed untrained anomaly detection methods for images with periodic texture, including spectral methods and low-rank decomposition methods (Cao et al., 2017; Hou and Zhang, 2007; Shi et al., 2021). However, most of these methods were developed for the fabrics industry and have limitations when applied to more complex periodic surfaces.

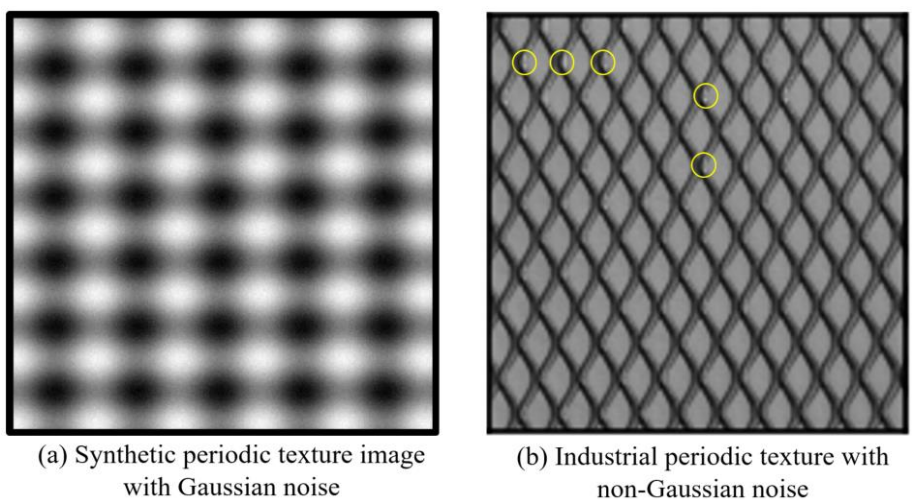

(a) Synthetic periodic texture image with Gaussian noise

(b) Industrial periodic texture with non-Gaussian noise

Figure 1. Periodic images with (a) Gaussian noise and (b) non-Gaussian noise

One significant limitation is that periodic pattern information cannot be exactly extracted, thereby hindering the estimation of pixel distribution and further anomaly detection. Moreover, due to the quality inspection environment, different noises will occur for precise anomaly detection. For example, Figure 1 shows the synthetic periodic image with Gaussian noise and a real industrial periodic image with non-Gaussian noise. The pattern marked in the circle of Figure 1(b) is caused by environmental lighting, which is usually misidentified as anomalies. The existing detection methods cannot deal with this challenge because they simply locate the anomalies without the estimation of pixel distribution.

To fill the above research gaps, we aim to develop an untrained anomaly detection methodology for real industrial images with periodic textures. This methodology will learn periodic patterns and handle textures with complicated noises. However, achieving this goal is challenging due to the following factors: (i) Accurately extracting periodic patterns from a single image is difficult, especially in the presence of anomalies and significant noise. (ii) The images of periodic surfaces may have varying periodic directions. They may not exhibit periodic patterns in either the horizontal or vertical direction, which makes it challenging to recognize the periodic patterns. (iii) Due to the randomness of manufacturing environments, some regions of the surfaces may not strictly satisfy the periodic assumption. This effect is especially pronounced in pixel regions with high gradients in real-world industrial images, thereby leading to false detection as anomalies. Here, the high gradients of pixels mean that the neighboring pixel values have a large difference.

To address the challenges mentioned above, we propose a novel approach for representing periodic patterns and apply it to untrained anomaly detection tasks for periodic images. The key contributions of our work are as follows:

- To address the first challenge, we propose a novel representation method for periodic patterns. Concretely, we introduce a novel self-representation defined by continuous parameters to represent an ideal periodic image. Subsequently, we propose a joint optimization framework named periodic-sparse decomposition (PSD), designed to simultaneously estimate the periodic background, sparse anomalies, and Gaussian noise.

- The PSD framework requires horizontal or vertical periodic directions. To accommodate PSD to images with diverse periodic directions, as stated in the second challenge, we propose an image rotation and reconstruction module. This module adjusts the image orientation to match the required periodic direction and reconstructs the reference image accordingly.
- For the real industrial images, we observe that the values of normal pixels violating the strict periodicity are associated with larger variabilities. Therefore, incorporating the variability of pixel values into anomaly scores can effectively alleviate the false detection issue. To this end, we propose a novel pixel-level anomaly strategy based on normalized distance to enhance the anomaly detection performance.

The remainder of this article is organized as follows: Section 2 reviews the related literature on anomaly detection methods for periodic images. Section 3 introduces the proposed periodic pattern representation, the PSD framework, and the overall procedure for processing real industrial images. Section 4 provides extensive case studies in numerical and real images to validate our method. Finally, this article is concluded in Section 5.

## 2. Literature Review

Anomaly detection methods for periodic images can be categorized based on their dependency on a training process and dataset into two distinct types: training-based methods and untrained methods.

### 2.1 *Training-based Anomaly Detection for Periodic Images*

Training-based methods aim to learn normal appearances and features, not limited to the periodicity, from the training dataset. In the training stage, features are extracted from training images by Gabor filtering or modern deep learning techniques in the literature, enabling anomaly detection in subsequent test images.

Gabor filtering-based techniques (Jia et al., 2017; Mak and Peng, 2008; Mak et al., 2009) utilize Gabor filters to capture fundamental features from anomaly-free images. These features are then used to design specific Gabor filters tailored for anomaly detection, capable of identifying anomalies against backgrounds with textures similar to the anomaly-free training images.

Deep learning methods, on the other hand, employ unsupervised visual anomaly detection models. These models distinguish anomalies by contrasting the features of normal and abnormal samples (Bergmann et al., 2020; Yu et al., 2021). Deep learning methods are not constrained by texture regularity, allowing them to handle both regular and irregular textures, with generally superior performance observed in regular textures.

The primary limitations of training-based methods stem from the need for training datasets and the process of model training. Acquiring and annotating datasets for different textures can be both time-consuming and expensive.

## 2.2 *Untrained Anomaly Detection for Periodic Images*

Untrained methods for detecting anomalies in periodic images encompass the following categories:

1) **Statistical Methods:** Statistical techniques employ the sliding window technique to capture patches of the input image, allowing for the analysis of the spatial feature distribution of gray values, including gray-level co-occurrence matrices (GLCM), autocorrelation analysis, and fractal dimension analysis. Pixels with statistical features of surrounding pixels that deviate from the distribution will be detected as anomalies. For example, Raheja et al. extracted textural features from fabric images using GLCM and implemented a sliding window technique for anomaly detection. This method involves the window moving across the entire image, computing textural energy from the GLCM of the fabric image. The energy values are then compared to a reference, with deviations beyond a threshold reported as anomalies (Raheja, Ajay, et al., 2013; Raheja, Kumar, et al., 2013). These statistical methods are adept at identifying anomalies in fabrics with simple and uniform weaves, such as plain and twill.

2) **Spectral Methods:** Fourier and wavelet transforms are widely applied for image anomaly detection. Hou and Zhang introduced an innovative object detection technique utilizing the log Fourier spectra of images, which proves effective in anomaly detection without necessitating prior background knowledge (Hou and Zhang, 2007). However, this method is sensitive to noise and requires a strict periodic background. Yang et al. developed a fabric anomaly detection method utilizing an adaptive wavelet-based feature extractor (Yang et al., 2002). In this method, the periodicity is attributed to the yarn fabric, while more general periodic patterns remain unexplored.

3) **Low-rank Decomposition-based Methods:** Low-rank decomposition techniques have become increasingly prevalent for anomaly detection on periodic images. These methods capitalize on the intrinsic low-rank property of periodic images, whether in feature space or within the original data itself.

- **Low-rank Decomposition of Original Image Matrix:** For most textile textures, including uniform and periodic textures, the matrix of pixel values of the input image can be considered as a low-rank matrix. The anomalies in periodic textures can be regarded as sparse matrices (Huangpeng et al., 2018). By leveraging the characteristics of low-rank background and sparse anomalies to estimate the low-rank and sparse components, low-rank decomposition methods are utilized for anomaly detection in textile images (Mo et al., 2021; Shi et al., 2019, 2021). However, the low-rank decomposition of the original image matrix fails to handle periodic textures with periodic patterns in varying directions (Tsai and Hsieh, 1999).

- **Low-rank Decomposition of Features:** As features capture more semantic information than individual pixels, the generalization of low-rankness into feature space introduces greater applicability to real images. Li et al. proposed an efficient second-order orientation-aware descriptor, denoted as GHOG, combining Gabor and histogram of oriented gradient (HOG) (Li et al., 2019). Based on the proposed GHOG, a low-rank decomposition model was developed for fabric anomaly detection. Furthermore, Cao et al. introduced a least-square regression model guided by prior knowledge to replace the popular nuclear norm for low-rank representation, improving the computational efficiency (Cao et al., 2017). One limitation of feature low-rank decomposition approaches is that they are patch-level anomaly detection, resulting in inaccurate shape descriptions of the anomalies and poor performance when dealing with small and subtle anomalies.

In summary, the literature still lacks an untrained anomaly detection method to deal with periodic images with concentrated noises. A viable approach to deal with this challenge is to extract the periodic pattern and estimate the pixel distribution. To fill the above research gap, we propose the PSD-based methodology as illustrated in Section 3.

## 3. PSD-based Methodology

The intrinsic representation for an ideal periodic signal can be defined by its fundamental period $T$, i.e., the interval between successive repetitive units. With a known $T$, one can infer the expected value of a query point $p$ by combining the in-phase points, whose distances to $p$ are integer multiples of $T$.

However, estimating $T$ from discrete data (such as time series and images) is a highly nonconvex problem and challenging to optimize (Liu et al., 2004; Quinn and Thomson, 1991; Rife and Boorstyn, 1974). To avoid the explicit estimation of $T$, we propose a novel self-representation of the periodic signal defined on a set of continuous parameters in Subsection 3.1. To account for the presence of sparse anomalies and random noise, we further develop an optimization framework named Periodic-Sparse Decomposition (PSD) in Subsection 3.2. Finally, we propose our untrained anomaly detection method for real cases with irregular periodic directions and the non-strict periodic assumption in Subsection 3.3.

## 3.1 *A Novel Periodic Pattern Representation Method*

We first discuss the self-representation model for 1D periodic signals and then generalize it to 2D images.

### 3.1.1 *Self-Representation for 1D Signals*

In this subsection, we begin by developing a formal description of the 1D periodic signal and relate it with the estimation of the period $T$ by summarizing a self-representation equation.

We denote $\boldsymbol{y} = [y_1 \ y_2 \ \cdots \ y_n] \in \mathbb{R}^n$ as an ideal periodic signal. Since $\boldsymbol{y}$ is ideal periodic, for $\forall i = 1,2,\cdots,n$, $y_i$ equals the weighted average of the values of the in-phase points ($y_j$, where $j = kT + i, k \in \mathbb{Z}$, and $j \in [1,n]$). That is, for $\forall i = 1,2,\cdots,n$,

$$y_i = \sum_{j=1}^{n} R_{i,j} y_j,$$

$$s.t. \ R_{i,j} \begin{cases} \geq 0, & j = kT + i, k \in \mathbb{Z} \\ = 0, & \text{others} \end{cases} ; \ \sum_{j=1}^{n} R_{i,j} = 1. \tag{1}$$

Denoting the matrix $\boldsymbol{R} \in \mathbb{R}^{n \times n}$ collects all $\{R_{i,j}\}$, we organize the above Eq. (1) into a matrix form, i.e., $\boldsymbol{y} = \boldsymbol{R}(T) \cdot \boldsymbol{y}$, which is a self-representation of $\boldsymbol{y}$ w.r.t. $T$. Notably, there is an infinite number of solutions for $\boldsymbol{R}$.

Our objective is to find an optimal $T$ that satisfies Eq. (1), which is challenging in practice as $T$ is required as a positive integer. Enumerating all possible $T$ is computationally intensive, especially when the task of periodic pattern learning is embedded into another top-level task, e.g., anomaly detection, see Section 3.2. To address it, we introduce an $n$-dimensional continuous vector, $\boldsymbol{\lambda} = [\lambda_0, \lambda_1, \ldots, \lambda_{n-1}]^T \in \mathbb{R}^n$, referred to as the periodic pattern vector. This vector is used to construct an $\boldsymbol{R}(\boldsymbol{\lambda})$ with the form $\boldsymbol{R}(\boldsymbol{\lambda}) = \boldsymbol{W}(\boldsymbol{\lambda}) \cdot \boldsymbol{S}(\boldsymbol{\lambda})$. In this expression, $\boldsymbol{S}(\boldsymbol{\lambda})$ is an $\mathbb{R}^{n \times n}$ matrix with elements $S_{i,j} = \lambda_{|i-j|}$, and $\boldsymbol{W}(\boldsymbol{\lambda})$ is an $\mathbb{R}^{n \times n}$ diagonal matrix where each diagonal entry $W_{ii} = \frac{1}{\sum_{j=1}^{n}|\lambda_{|i-j|}|}$.

The following Proposition 1 indicates that if $\boldsymbol{\lambda}$ meets certain conditions, we can directly derive a solution of $\boldsymbol{R}$ in Eq. (1). Furthermore, we can relax the requirement of $\boldsymbol{\lambda}$ by simply finding a sparse $\boldsymbol{\lambda}$. Subsequently, since $\boldsymbol{\lambda}$ is defined on $n$-dimensional continuous real space, the optimization is more feasible than directly estimating $T$, which can be dealt with by the gradient descent-based algorithms.

***Proposition 1***. If the $l^{th}$ entry $\lambda_l$ of the periodic pattern vector $\boldsymbol{\lambda}$ is only non-zero for $l = kT$, $k \in \mathbb{N}$, the matrix $\boldsymbol{R}(\boldsymbol{\lambda}) = \boldsymbol{W}(\boldsymbol{\lambda}) \cdot \boldsymbol{S}(\boldsymbol{\lambda})$ will satisfy the constraints in Eq. (1). Additionally, $\boldsymbol{R}(\boldsymbol{\lambda})$ satisfies the self-representation condition stipulated by $\boldsymbol{y} = \boldsymbol{R}(\boldsymbol{\lambda}) \cdot \boldsymbol{y}$.

***Discussion of Proposition 1***. Our approach involves a strategic relaxation of the first constraint of $\boldsymbol{R}$ as specified in Eq. (1), i.e., the periodicity requirement, through the use of the symmetric Toeplitz matrix $\boldsymbol{S}(\boldsymbol{\lambda})$. The matrix $\boldsymbol{S}(\boldsymbol{\lambda})$ is structured such that $S_{i,j}$ is non-zero only when $|i - j| = kT$, consistent with the periodicity constraint of Eq. (1). In addition to $\boldsymbol{S}(\boldsymbol{\lambda})$, we incorporate the matrix $\boldsymbol{W}(\boldsymbol{\lambda})$ as a normalization factor, which effectively embodies the second constraint of Eq. (1). Hence, once $\boldsymbol{\lambda}$ has the property in Proposition 1, it can be easily checked that Eq. (1) holds true.

### 3.1.2 *Self-Representation for 2D Images*

Next, we discuss how to generalize the above proposition for the ideally periodic image $\boldsymbol{Y} \in \mathbb{R}^{n \times n}$, which presents periodicity in both the horizontal and vertical directions. We denote $\boldsymbol{\lambda_1}$ and $\boldsymbol{\lambda_2}$ as periodic pattern vectors along vertical and horizontal directions respectively. From Proposition 1, we have

$$\boldsymbol{Y} = \boldsymbol{W}(\boldsymbol{\lambda_1}) \cdot \boldsymbol{S}(\boldsymbol{\lambda_1}) \cdot \boldsymbol{Y}, \qquad (2)$$

$$\boldsymbol{Y}^T = \boldsymbol{W}(\boldsymbol{\lambda_2}) \cdot \boldsymbol{S}(\boldsymbol{\lambda_2}) \cdot \boldsymbol{Y}^T, \qquad (3)$$

where Eqs. (2) and (3) indicate the vertical and horizontal periodicity, respectively. Combining Eqs. (2) and (3), we obtain the joint self-representation of $\boldsymbol{Y}$ as

$$\boldsymbol{Y} = \boldsymbol{W}(\boldsymbol{\lambda_1}) \cdot \boldsymbol{S}(\boldsymbol{\lambda_1}) \cdot [\boldsymbol{W}(\boldsymbol{\lambda_2}) \cdot \boldsymbol{S}(\boldsymbol{\lambda_2}) \cdot \boldsymbol{Y}^T]^T. \qquad (4)$$

## 3.2 *Periodic-Sparse Decomposition (PSD)*

In real-world industrial data, noise is inevitable, and anomalies may occur in some defective products. Anomalies are defined as faults whose value differ from the expected value of the periodic background significantly (Yan et al., 2017). To deal with this case, we first decompose the input signal to periodic background, sparse anomalous regions, and Gaussian noise. Then, a novel optimization framework, namely PSD, is presented to estimate the above components. We first develop the model for 1D signals and then generalize it to 2D images.

### 3.2.1 *Periodic Signal Decomposition Optimization Framework*

A 1D periodic signal can be decomposed into three parts: $\boldsymbol{y} = \boldsymbol{y}^* + \boldsymbol{a} + \boldsymbol{e}$. $\boldsymbol{y}^* \in \mathbb{R}^n$ represents the ideal periodic signal without anomalies or noise, $\boldsymbol{a} \in \mathbb{R}^n$ represents the sparse anomalies, and $\boldsymbol{e} \in \mathbb{R}^n$ denotes the random noise.

Based on the above decomposition, we can estimate $\boldsymbol{\lambda}$, $\boldsymbol{a}$, and $\boldsymbol{e}$ by solving the following constrained problem

$$\min_{\boldsymbol{\lambda},\boldsymbol{a},\boldsymbol{e}} \|\boldsymbol{y}^* - \boldsymbol{R}(\boldsymbol{\lambda}) \cdot \boldsymbol{y}^*\|_2^2 + \beta_1 \|\boldsymbol{a}\|_1 + \beta_2 \|\boldsymbol{e}\|_2^2,$$

$$s.t.\ \boldsymbol{y} = \boldsymbol{y}^* + \boldsymbol{a} + \boldsymbol{e}. \qquad (5)$$

The above Problem (5) means that we aim to promote the periodicity of $\boldsymbol{y}^*$ according to our Proposition 1, while encouraging the sparsity of $\boldsymbol{a}$ and controlling the magnitude of $\boldsymbol{e}$. It is challenging to solve Problem (5) due to its non-convex nature. On the other hand, we can observe that Problem (5) becomes convex w.r.t. $\boldsymbol{a}$ and $\boldsymbol{e}$ for a fixed $\boldsymbol{\lambda}$, which can be easily handled. Therefore, we aim to solve Problem (5) by alternating update $\boldsymbol{\lambda}$, $\boldsymbol{a}$, and $\boldsymbol{e}$.

However, for the fixed $\boldsymbol{a}$, and $\boldsymbol{e}$, the term $\|\boldsymbol{y}^* - \boldsymbol{R}(\boldsymbol{\lambda}) \cdot \boldsymbol{y}^*\|_2^2$ in (5) is still non-convex w.r.t. $\boldsymbol{\lambda}$. To simplify the optimization and concurrently ensure the accurate estimation of $\boldsymbol{\lambda}$, we further modify the problem (5) by incorporating convex approximation and additional regularizations as follows.

- We approximate $\|y^* - R(\lambda) \cdot y^*\|_2^2$ with $\|y^* - S(\lambda) \cdot y^*\|_2^2$. The above approximation indicates that $W(\lambda)$ is required to be close to the identity matrix $I_n$. To achieve it, we introduce the penalty $\sum_{i=1}^{n}(\mathbf{1}_n^T \cdot |S_i| - 1)^2$ with a tuning parameter $\rho_1$. This strategy enables a convex objective function w.r.t. $\lambda$.

- Proposition 1 shows that the optimal solution of $\lambda$ is sparse. Therefore, we use $L_1$ penalty $\|\lambda\|_1$ with tuning parameter $\rho_2$ to promote the sparsity of $\lambda$ to get a better solution. To verify the effect of $\rho_2\|\lambda\|_1$, we test PSD under different $\rho_2$ values, and the results are shown in Appendix A.

- To avoid obtaining the trivial solution such as $\lambda = [1 \quad 0 \quad \cdots \quad 0]^T$, we restrict the first $p$ components of $\lambda$ to zero. Due to the penalty $\sum_{i=1}^{n}(\mathbf{1}_n^T \cdot |S_i| - 1)^2$, when the first $p$ components of $\lambda$ are 0, the last $p$ components of $\lambda$ will ultimately be optimized to 0. Therefore, we also set the last $p$ components to 0 directly.

Considering the above approximations and penalty terms, the objective function for optimizing $\lambda$ with fixed $a$ and $e$ is rewritten as:

$$\min_{\lambda} \mathcal{L}_1(\lambda) = \|(y - a - e) - S(\lambda) \cdot (y - a - e)\|_2^2 + \rho_1 \sum_{i=1}^{n}(\mathbf{1}_n^T \cdot |S_i| - 1)^2 + \rho_2\|\lambda\|_1,$$

$$s.t. \lambda_i = 0, \ i = 0,1,\cdots,p-1, n-p,\cdots,n-1. \tag{6}$$

The objective function for optimizing $a$ with a fixed $\lambda$ and $e$ is rewritten as Problem (7):

$$\min_{a} \mathcal{L}_2(a) = \|(y - a - e) - W(\lambda) \cdot S(\lambda) \cdot (y - a - e)\|_2^2 + \beta_1\|a\|_1. \tag{7}$$

The objective function for optimizing $e$ with fixed $\lambda$ and $a$ is rewritten as the Problem (8).

$$\min_{e} \mathcal{L}_3(e) = \|(y - a - e) - W(\lambda) \cdot S(\lambda) \cdot (y - a - e)\|_2^2 + \beta_2\|e\|_2^2. \tag{8}$$

The adaptive moment estimation (Adam) optimization algorithm is well-suited for solving convex problems due to its efficiency and adaptive learning rates. We solve Problems (6) and (7) using the Adam algorithm (Kingma and Ba, 2017) implemented by PyTorch (Paszke et al., 2017) in Python. Problem (8) is a ridge regression problem with an analytical solution $\hat{e} = (X^TX + \beta_2 I_n)^{-1}X^T[X(y - a)]$, where $X = I_n - W(\lambda) \cdot S(\lambda)$. The algorithm for optimizing $\lambda$, $a$, and $e$ is summarized in Algorithm 1.

### 3.2.2 *Periodic Image Decomposition Optimization Framework*

A 2D periodic image can be decomposed into three parts: $Y = Y^* + A + E$. $Y^* \in \mathbb{R}^{n \times n}$ represents the ideal periodic image without anomalies or noise and presents periodicity in both horizontal and vertical directions, $A \in \mathbb{R}^{n \times n}$ represents the sparse anomalies, and $E \in \mathbb{R}^{n \times n}$ denotes the random noise. For two-dimensional image decomposition, we also alternatively optimize periodic pattern vectors $\lambda_1, \lambda_2$, anomalies $A$, and noises $E$. Similar to Problems (6-8) for a one-dimensional signal, the objective functions for a two-dimensional image are proposed as follows.

The objective function to update $\lambda_1$ with fixed $\lambda_2, A$, and $E$ is rewritten as the Problem (10).

$$\min_{\lambda_1} \mathcal{L}_4(\lambda_1) = \|(Y - A - E) - S(\lambda_1) \cdot (Y - A - E)\|_F^2 + \rho_1 \sum_{i=1}^n \left(\mathbf{1}_n^T \cdot |S_i(\lambda_1)| - 1\right)^2$$

$$+ \rho_2 \|\lambda_1\|_1,$$

$$s.t. \lambda_{1,i} = 0, \ i = 0, 1, \cdots, p - 1, n - p, \cdots, n - 1. \tag{10}$$

The objective function to update $\lambda_2$ with fixed $\lambda_1, A$, and $E$ is rewritten as the Problem (11).

$$\min_{\lambda_2} \mathcal{L}_5(\lambda_2) = \|(Y - A - E) - [S(\lambda_2) \cdot (Y - A - E)^T]^T\|_F^2 + \rho_1 \sum_{i=1}^n \left(\mathbf{1}_n^T \cdot |S_i(\lambda_2)| - 1\right)^2$$

$$+ \rho_2 \|\lambda_2\|_1,$$

$$s.t. \lambda_{2,i} = 0, \ i = 0, 1, \cdots, p - 1, n - p, \cdots, n - 1. \tag{11}$$

The objective function to update $A$ with fixed $\lambda_1, \lambda_2$, and $E$ is rewritten as the Problem (12).

$$\min_A \mathcal{L}_6(A) = \|(Y - A - E) - W(\lambda_1) \cdot S(\lambda_1) \cdot (Y - A - E) \cdot [W(\lambda_2) \cdot S(\lambda_2)]^T\|_F^2$$

$$+ \beta_1 \|A\|_1. \tag{12}$$

Here, $\|A\|_1 = \sum_{ij}|A_{ij}|$ is the entrywise $L_1$ norm of the $A$ matrix.

**Algorithm 1. Optimization algorithm for PSD for One-dimensional Signal**

**input** $y, \epsilon, K$
**initialize** $\lambda, a, e$
$t = 0$
**while** $\|\lambda^t - \lambda^{t-1}\|_\infty > \epsilon$ **or** $t = 0$ **do**
    $t = t + 1, k = 0$
    **while** $\|\lambda^k - \lambda^{k-1}\|_\infty > \epsilon$ **or** $k = 0$ **do**
        $k = k + 1$
        Update $\lambda = \text{Adam}(\lambda, \mathcal{L}_1(\lambda))$         ▷$\mathcal{L}_1(\lambda)$ refers to Problem (6)
    **end while**
    **for** $k = 1$ *to* $K$ **do**
        Update $a = \text{Adam}(a, \mathcal{L}_2(a))$         ▷$\mathcal{L}_2(a)$ refers to Problem (7)
    **end for**
    **for** $k = 1$ *to* $K$ **do**
        Update $e = (X^T X + \beta_2 I_n)^{-1} X^T [X(y - a)], \; X = I_n - W(\lambda) \cdot S(\lambda)$
    **end for**
**end while**
$y^* = y - a - e$         (9)
**output** $\lambda, a, y^*$

**Algorithm 2. Optimization Algorithm for PSD for Two-dimensional Images**

**input** $Y, \epsilon, K$
**initialize** $\lambda_1, \lambda_2, A$,
$t = 0$
**while** $\|\lambda_1^t - \lambda_1^{t-1}\|_\infty > \epsilon$ **or** $\|\lambda_2^t - \lambda_2^{t-1}\|_\infty > \epsilon$ **do**
    $t = t + 1, k = 0$
    **while** $\|\lambda_1^k - \lambda_1^{k-1}\|_\infty > \epsilon$ **or** $\|\lambda_2^k - \lambda_1^{k-1}\|_\infty > \epsilon$ **or** $k = 0$ **do**
        $k = k + 1$
        Update $\lambda_1 = \text{Adam}(\lambda_1, \mathcal{L}_4(\lambda_1))$     ▷$\mathcal{L}_4(\lambda_1)$ refers to Problem (10)
        Update $\lambda_2 = \text{Adam}(\lambda_2, \mathcal{L}_5(\lambda_2))$     ▷$\mathcal{L}_5(\lambda_2)$ refers to Problem (11)
    **end while**
    **for** $k = 1$ **to** $K$ **do**
        Update $A = \text{Adam}(A, \mathcal{L}_6(A))$     ▷$\mathcal{L}_6(A)$ refers to Problem (12)
    **end for**
    **for** $k = 1$ **to** $K$ **do**
        Update $E = \text{Adam}(E, \mathcal{L}_7(E))$     ▷$\mathcal{L}_7(E)$ refers to Problem (13)
    **end for**
**end while**
$Y^* = Y - A - E$         (14)
**output** $\lambda_1, \lambda_2, A, Y^*$

The objective function to update $E$ with fixed $\lambda_1, \lambda_2$, and $A$ is rewritten as the Problem (13).

$$\min_{E} \mathcal{L}_7(E) = \|(Y - A - E) - W(\lambda_1) \cdot S(\lambda_1) \cdot (Y - A - E) \cdot [W(\lambda_2) \cdot S(\lambda_2)]^T\|_F^2$$

$$+\beta_2 \|E\|_2^2. \tag{13}$$

Although Problem (13) has an analytical solution. However, this is a large-scale ridge regression problem with a large coefficient matrix, i.e., the dimension of $n^2 \times n^2$, where computation of analytical solution is expensive. To speed up, we solve it by using stochastic gradient descent algorithms (Zhang, 2004) via the Adam algorithm. The algorithm for optimizing $\lambda_1, \lambda_2, A$, and $E$ is summarized in Algorithm 2.

### 3.3 *Untrained Anomaly Detection for Real Industrial Periodic Images*

The PSD framework requires that images have regular periodic patterns in the horizontal or vertical direction. However, for real-world applications, the above requirement may not be satisfied, as illustrated in Figure 2. Furthermore, certain regions may not strictly satisfy the periodicity assumption due to manufacturing randomness and non-confronted measuring perspectives. These regions are prone to be misdetected as anomalies in PSD. To address the above two problems, we propose a novel framework for anomaly detection of real periodic images, including (i) Reference image construction in Section 3.3.1, which recovers the ideal periodic image for the inclined input image. (ii) Pixel-level anomaly scoring in Section 3.3.2, which enables accurate anomaly labeling for the case of non-strict periodicity. The overall procedure is illustrated in Figure 2.

#### 3.3.1 *Reference Image Reconstruction*

The construction of the reference image contains four steps: image rotation, PSD, image expansion, and image reverse rotation.

The first step rotates the image to align its periodic direction with either the horizontal or vertical axis. The periodic direction of an image can be determined by analyzing its Fourier transform spectrum (Tsai and Hsieh, 1999). The output of this step is a rotated image in which the periodic directions satisfy the requirement for PSD. The "Rotated Image" in Figure 2 rotates the original image, with the periodic directions and lengths indicated by arrows.

In the second step, PSD is applied to the rotated image, providing the periodic pattern vectors $\lambda_1$, $\lambda_2$, and the rotated reference image $Y^*$. The first step decreases the size of the rotated image. Therefore, we need to expand the rotated reference image to ensure sufficient pixels, which can be implemented by our learned periodic pattern vectors. We will discuss the details in Appendix B.

Finally, the reference image that remains the same size as the input image is constructed by reversely rotating the expanded-rotated reference image to get the final reference image, as demonstrated in Figure 2.

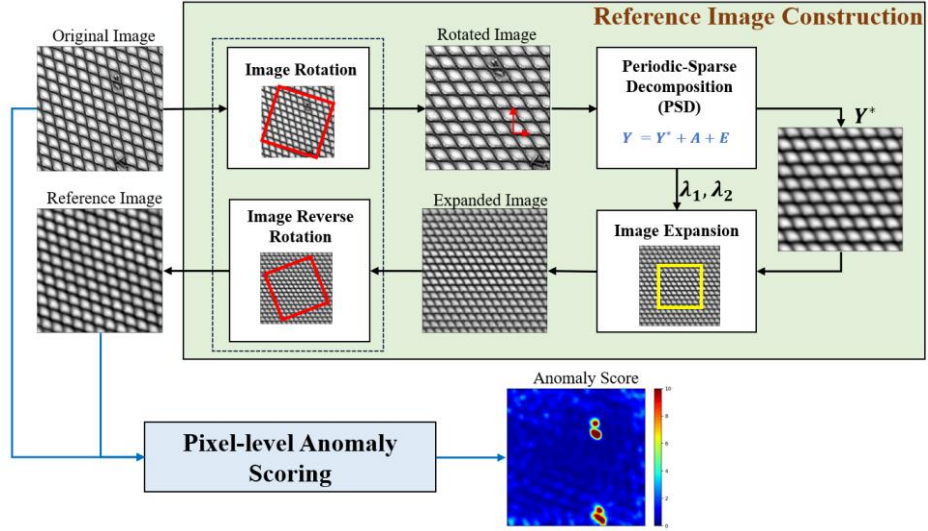

Figure 2. Overview of the untrained anomaly detection method

### 3.3.2 *Pixel-Level Anomaly Scoring*

After obtaining the reference image, a straightforward method for anomaly detection is to calculate the difference between the reference image and the original image. However, the pixel value differences are large even in some regions without anomalies, as illustrated in the region covered by the rectangles of Figure 3(a). The reason is that the pixel values of these regions are not strictly periodic due to inspection environments. To improve the detection accuracy, we propose a novel pixel-level anomaly scoring strategy as follows.

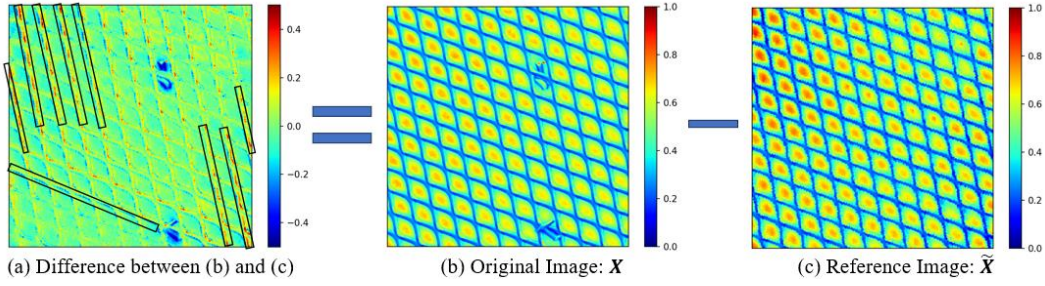

(a) Difference between (b) and (c)   (b) Original Image: $X$   (c) Reference Image: $\widetilde{X}$

Figure 3. Case of (a) the difference between (b) the original image and (c) the reference image

The anomaly-free regions that violate the strict periodic assumption also exhibit certain periodic structures, as shown in Figure 3, which appear at the edges of the grid. In these areas, a wide range of values can be considered "normal" pixel values. This means that the variability of the pixel value at a specific location should be incorporated into anomaly scoring. For a target pixel, we can find its counterparts in other locations by the periodicity of the input image, whose values follow the same distribution. Subsequently, with these counterparts, we can obtain the desired variability by calculating the variance of the associated distribution. An overview of the method is presented in Figure 4. The process of evaluating the anomaly score for each pixel involves two steps: (i) Calculation of the normalized pixel distances, and (ii). Convolution with a Gaussian kernel.

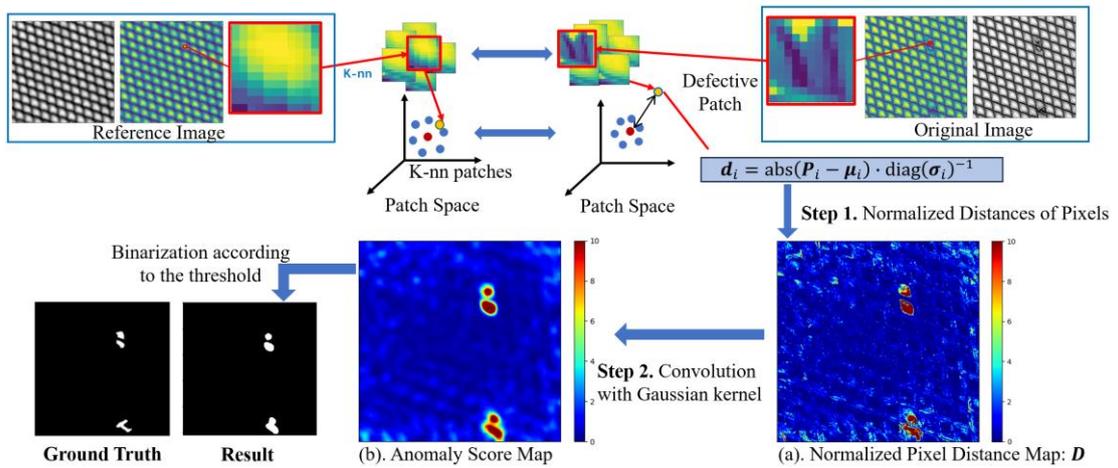

Figure 4. Pixel-level anomaly scoring

To achieve the above purpose, we utilize the image patch representation. Specifically, we can segment both the reference image and the original image into $N$ square patches, i.e., $\mathcal{P} = \{P_i \in \mathbb{R}^M, i = 1, 2, \cdots N\}$ and $\tilde{\mathcal{P}} = \{\tilde{P}_i \in \mathbb{R}^M, i = 1, 2, \cdots N\}$, respectively. $M$ denotes the number of pixels in each patch.

Using the patch set of the reference image, we search $K$ nearest similar structured patches for each patch. That is, for the $i^{th}$ patch $\tilde{P}_i \in \tilde{\mathcal{P}}$, we query $K$ nearest patches and denote the set of their indices as $\mathcal{J}_i = \{j_k \in \{1, 2, \cdots N\}, k = 1, 2, \cdots, K\}$. We then locate the corresponding patches in the original image for these $K$ patches $\mathcal{NN}_i = \{P_j \in \mathcal{P}, j \in \mathcal{J}_i\}$. We assume that each entry of a patch in $\mathcal{NN}_i$ follows a Gaussian distribution. Then, we concatenate the means and standard deviations of patch rows into vectors $\mu_i \in \mathbb{R}^M$ and $\sigma_i \in \mathbb{R}^M$. Therefore, the anomaly scores assigned to the pixels covered by $P_i$ are defined by the normalized distances as:

$$d_i = \text{abs}(P_i - \mu_i) \cdot \text{diag}(\sigma_i)^{-1}, \tag{15}$$

Where $\text{abs}(P_i - \mu_i)$ takes the absolute value of the elements of $P_i - \mu_i$, and $\text{diag}(\sigma_i)^{-1}$ denotes the inverse of the diagonal matrix filled with the values in $\sigma_i$. The distance measured here reflects the number of standard deviations by which the pixel value deviates from the mean. Eq. (15) assigns smaller anomaly scores to pixels with greater variance. This property enables us to reduce the false detection of the edges of the grid (within black rectangles) in Figure 3. An example of the obtained normalized distance map is visualized in Figure 4(a).

Finally, an anomaly should cover a certain number of pixels. Therefore, to eliminate the isolated pixels with large normalized distances, we aggregate the scores within the neighborhoods. This can be achieved by convolving the distance map with a Gaussian kernel to obtain the final score.

## 4. Case Study

In this section, we generate synthetic images with noise and anomalies to evaluate the performance of PSD. Furthermore, we utilize the "Grid" dataset of real product images from the MVTec anomaly detection dataset to compare the performance of our proposed untrained anomaly detection method with benchmark methods. The MVTec dataset is widely used as a benchmark for evaluating industrial anomaly detection methods (Bergmann et al., 2021).

### 4.1 *Results on Numerical Images*

To evaluate the performance of PSD, we generate 30 images with sparse anomalies and random noise. The size of the images is $n \times n = 256 \times 256$. The periodic background of the image is generated using the function $Y^*_{i,j} = \sin\frac{2\pi i}{T_1} + \cos\frac{2\pi j}{T_2}$, where $T_1$ and $T_2$ are samples from uniform distribution U(50,80), so that the row vectors and column vectors of image matrix consist of 3-5 periodic repetitive units.

We employ the PSD technique to process the images, thereby extracting estimates of the periodic background $Y^*$, the noises $E$, and the anomalies $A$. The anomaly score is meticulously defined as $Sc = \frac{\text{abs}(A)}{\|A\|_F/n}$, where abs($A$) denotes the absolute value of the entries in the pixel matrix $A$. $\|A\|_F/n$ is the normalization factor derived from the root mean square of the entries in the pixel matrix $A$. This normalization ensures that the anomaly score is scaled appropriately for comparison. For example, in this experiment, a binary anomaly mask is calculated by applying the criterion $Sc_{ij} > 3.0$, indicating that pixels with anomaly scores exceeding this threshold are considered anomalies.

Figure 5 presents the analytical outcomes of a sample image, with the left panel depicting the original test image. The first row of Figure 5 sequentially displays the estimated $Y^*$, the ground truth noises $E$, the anomalies $A$, and the ground truth annotations $GT$. The second row of images illustrates our algorithm's estimations for $Y^*$, $E$, and $A$ as well as detection result. Additionally, the anomaly score image $Sc$ is positioned at the bottom right corner.

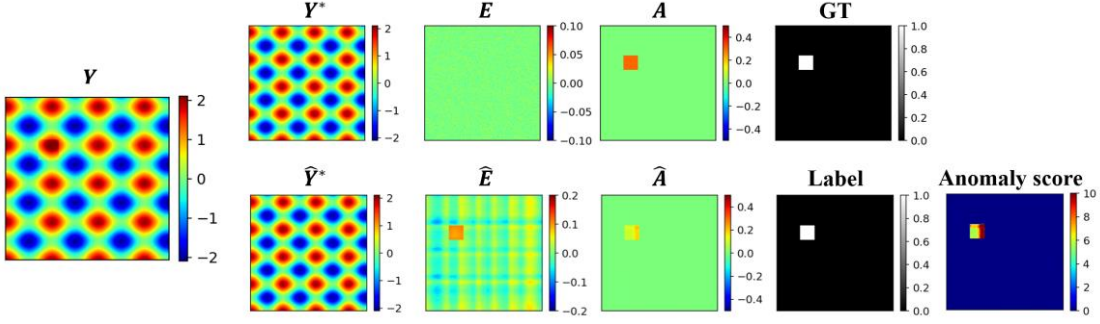

Figure 5. Synthetic image for illustration

We denote $FN$ and $TP$ as the counts of anomaly pixels that are incorrectly and correctly identified, respectively. Additionally, $FP$ and $TN$ represent the counts of normal pixels that are mistakenly and correctly classified, respectively. To assess the performance of our PSD method, we use the average of the following pixel-level anomaly detection indices (Tao and Du, 2023):

- The False Omission Rate (FOR), i.e., $FOR = \frac{FP}{TP+FP}$, indicates the system's capacity to avoid falsely classifying normal factors as anomalies.

- The False Negative Rate (FNR), i.e. $FNR = \frac{FN}{TP+FN}$, evaluates the system's capability to accurately detect anomalies.

- The Balanced Accuracy (BA), i.e., $BA = \frac{TP}{2(TP+FN)} + \frac{TN}{2(TN+FP)}$, follows the conventions of binary classification that measures the overall accuracy.

- The Dice coefficient (DICE), i.e., $DICE = \frac{2TP}{FP+FN+2TP}$, quantifies the overlap between the detected anomalies and the actual anomalies. Notably, lower values of FOR and FNR suggest superior performance, while higher values of BA and DICE indicate better alignment with the ground truth.

The average values of FOR, FNR, BA, and DICE are presented in Table 1 and depicted through box plots in Figure 6. From these results, it can be seen that our PSD performs well in detecting anomalies in ideal periodic images with horizontal or vertical periodic directions. It achieves remarkably low FOR and FNR values of 0.0011 and 0.0359, respectively, along with high BA and DICE scores of 0.9902 and 0.9800, respectively.

Table 1. Mean of FOR, FNR, BA, and DICE of detection results on 30 samples

| Index | FOR | FNR | BA | DICE |
|---|---|---|---|---|
| **Mean** | 0.0011 | 0.0359 | 0.9902 | 0.9800 |

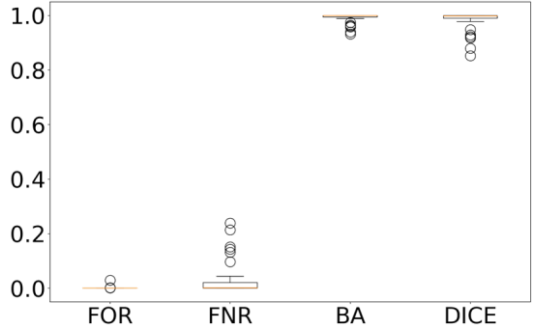

Figure 6. Box-plot of FPR, FNR, and DICE of 30 anomaly detection results

## 4.2 *Results on Real Images*

In this subsection, we present real case studies to compare the performance of our untrained anomaly detection method with that of other untrained methods using the Grid dataset.

The Grid dataset consists of a training dataset and a test dataset. The images in the training dataset of the Grid dataset are from good products. Notably, our method does not use the training dataset as the proposed PSD methodology is an untrained method. The test dataset of the Grid dataset includes good images and five categories of images of defective products, namely, "bent", "broken", "glue", "metal contamination", and "thread" anomalies. A statistical overview of the test set is provided in Table 2, and example images of each category are shown in Figure 7. The experiments were conducted on a computer with an Intel i7-13700H processor (2.40 GHz), 16.0 GB of RAM, and an NVIDIA RTX 4060 graphics card. The algorithm was implemented in Python with PyTorch (Paszke et al., 2017). The average processing time per image is approximately 5.4 seconds, and the total processing time for all 78 images in the test dataset is approximately 420 seconds.

Table 2. Testing samples of the grid dataset

| Category | good | bent | broken | glue | metal contamination | Thread |
|---|---|---|---|---|---|---|
| Number of Images | 21 | 12 | 12 | 11 | 11 | 11 |

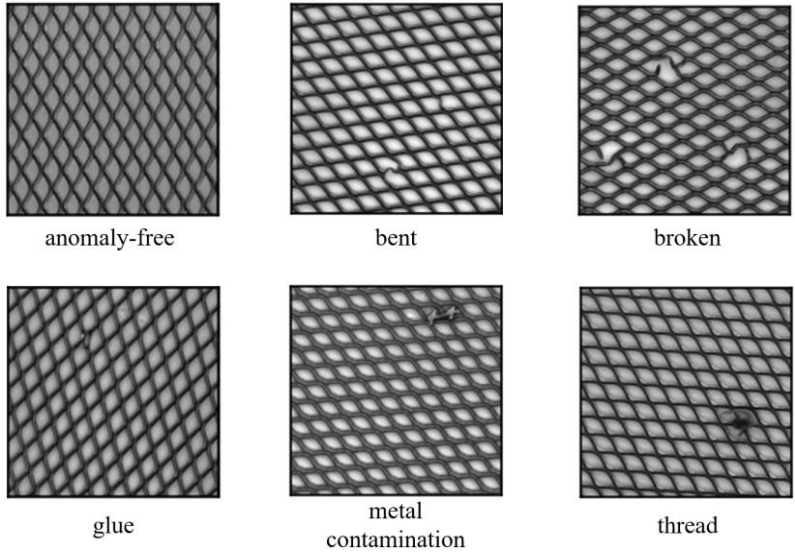

Figure 7. Example images of each category in the Grid dataset.

### 4.2.1 *Results of Our PSD Method*

Figure 8 presents a comparative visualization of our method, including 1 anomaly-free image and 5 defective images from distinct categories. The initial five columns, progressing from left to right, display the original images, the rotated sample images, the rotated reference images, the reference images, and anomaly score maps. To quantitatively assess the detection results against the ground truths, we used a threshold of 2.0 for identifying anomalies.

Notably, the testing Grid dataset presents distinct challenges not encountered in the detection of synthetic images, such as varying periodic directions, non-Gaussian noise, and diverse anomaly types. For instance, 'glue' anomalies, being relatively minor and subtle compared to other types of anomalies, are particularly susceptible to misclassification as normal pixels. Our method adeptly surmounts these obstacles, as evidenced by the representative results depicted in Figure 8. These findings highlight the method's robustness across various anomaly types and its overall effectiveness in real-world anomaly detection scenarios.

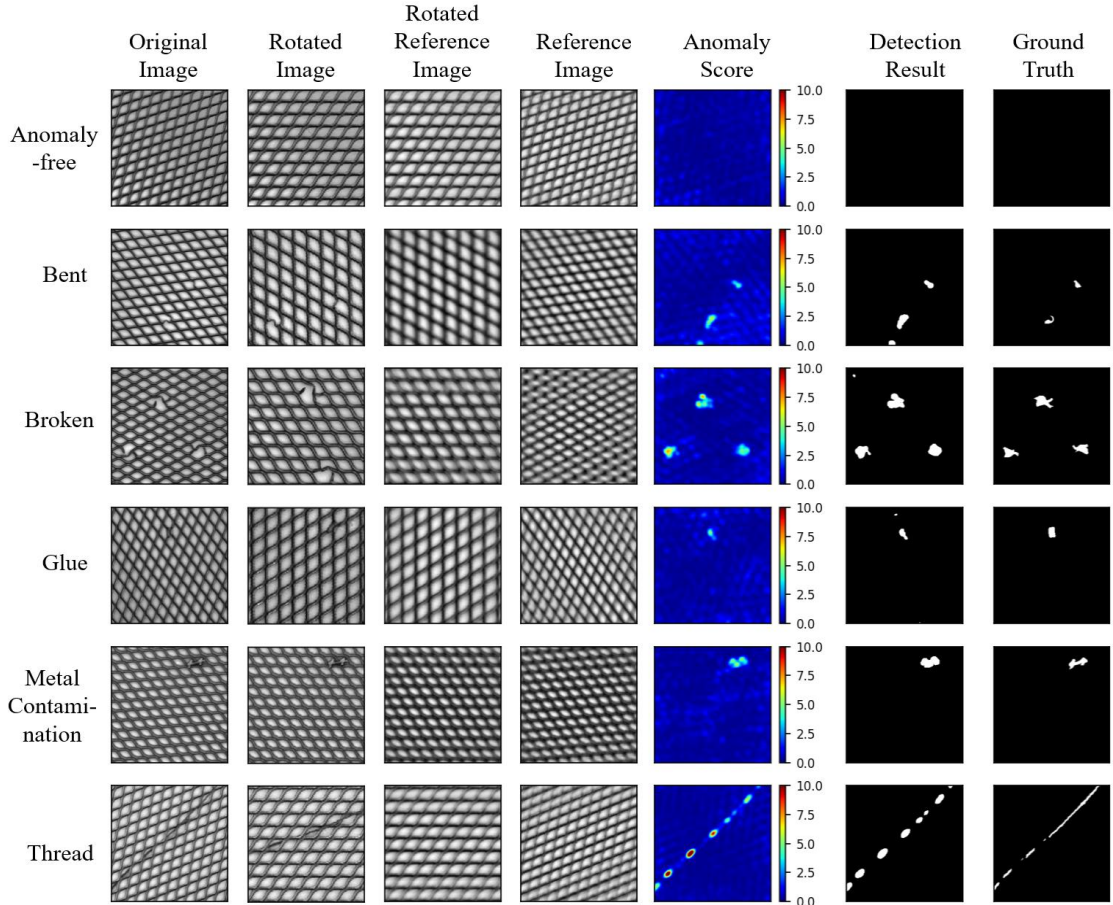

Figure 8. Visualization of output images of testing samples.

### 4.2.2 *Comparison with Existing Untrained Methods*

In this subsection, we compare our method with three untrained anomaly detection methods for regular image anomaly detection. The compared untrained methods include a feature low-rank decomposition method PG-LSR (Cao et al., 2017), a Spectral-Residual (SR) method (Hou and Zhang, 2007), and an image matrix low-rank decomposition method P-NLR (Shi et al., 2021).

Table 3 reports comprehensive evaluation indices to assess both our proposed method and benchmark untrained methods, where the best results are highlighted in bold. The FOR of our method is not quite as good as that of SR. The reason is that we adopt a progressive threshold such that the normal pixels near the anomalies will be mislabeled, as shown in the "glue" anomaly of Figure 9. Nevertheless, the comprehensive results demonstrate that our approach can strike the best balance between precision and recall, as illustrated by our superior performance in all 4 metrics as a whole. In addition, for each category of anomaly, our method consistently outperforms benchmark methods, showing its robustness and efficacy in real-world applications.

**Table 3.** Numerical results of different untrained methods

| Anomaly Type | Method | FOR (%) | FNR (%) | BA (%) | DICE (%) |
|---|---|---|---|---|---|
| all | PG-LSR | 80.17 | 77.10 | 64.07 | 18.37 |
|  | Spec-Res | **44.06** | 67.68 | 72.64 | 33.06 |
|  | P-NLR | 73.62 | 96.77 | 52.79 | 4.44 |
|  | Ours | 52.54 | **33.08** | **87.45** | **49.07** |
| bent | PG-LSR | 89.16 | 87.71 | 59.37 | 9.75 |
|  | Spec-Res | **20.06** | 73.14 | 69.63 | 36.26 |
|  | P-NLR | 86.71 | 98.50 | 51.32 | 2.21 |
|  | Ours | 54.78 | **29.99** | **89.53** | **51.04** |
| broken | PG-LSR | 88.99 | 83.94 | 60.70 | 10.81 |
|  | Spec-Res | **42.82** | 59.15 | 77.99 | 43.17 |
|  | P-NLR | 78.64 | 97.71 | 52.08 | 3.75 |
|  | Ours | 63.55 | **23.66** | **91.53** | **48.36** |
| glue | PG-LSR | 83.85 | 75.23 | 62.57 | 15.61 |
|  | Spec-Res | 44.21 | 73.65 | 68.77 | 22.77 |
|  | P-NLR | 59.95 | 98.08 | 51.64 | 2.89 |
|  | Ours | **38.95** | **71.99** | **69.91** | **35.54** |
| metal contamination | PG-LSR | 81.63 | 66.89 | 68.68 | 23.09 |
|  | Spec-Res | **41.85** | 57.55 | 78.27 | 38.58 |
|  | P-NLR | 93.88 | 94.79 | 54.24 | 4.46 |
|  | Ours | 61.69 | **7.19** | **97.75** | **53.02** |
| thread | PG-LSR | 57.21 | 71.73 | 69.02 | 32.60 |
|  | Spec-Res | 71.34 | 74.90 | 68.52 | 24.55 |
|  | P-NLR | 48.89 | 94.76 | 54.65 | 8.86 |
|  | Ours | **43.74** | **32.54** | **88.55** | **57.37** |

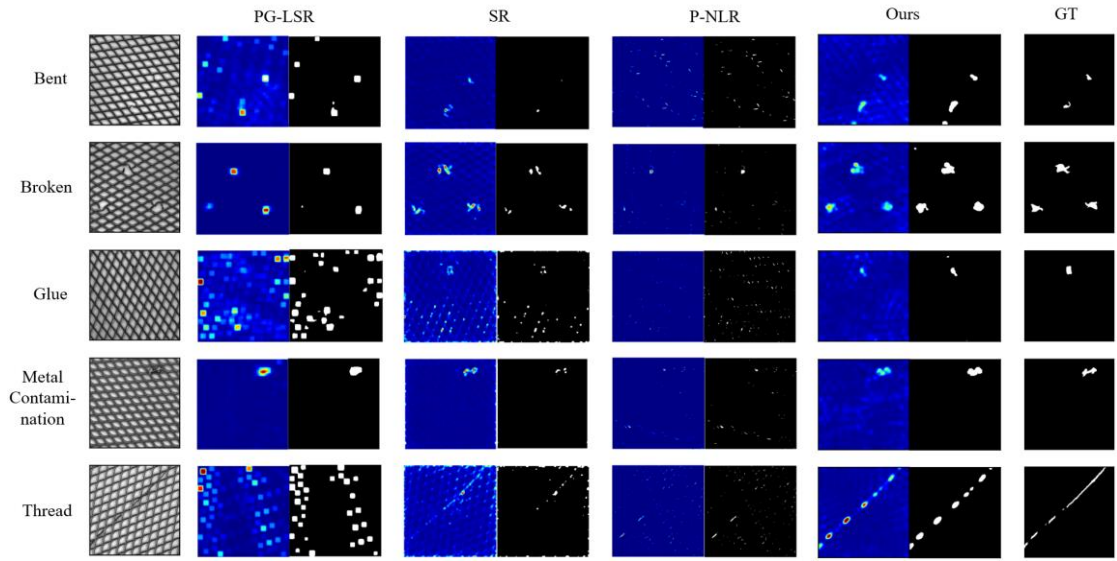

Figure 9. Anomaly localization results of untrained methods

## 5. Conclusion

Periodic image anomaly detection is important in various industrial applications. This paper addressed the challenges of learning periodic patterns in the presence of noise and anomalies. Specifically, we proposed a novel representation for images with periodic patterns. Additionally, we present a joint optimization framework named periodic-sparse decomposition (PSD) that takes both noise and anomalies into consideration to learn the proposed periodic pattern representation of images. To deal with the real industrial images with varying periodic directions, we designed an image rotation and reconstruction module. An additional anomaly scoring strategy was also proposed to alleviate the issue of non-strict periodicity. Both simulation and real case studies validate the accuracy and robustness of the proposed periodic pattern representation and untrained anomaly detection method, underscoring their potential to address the challenges of periodic surface anomaly detection.

## Code Availability

The code will be released upon publication.

# Acknowledgments

This work is supported by:

- The National Natural Science Foundation of China (No. 72371219).
- The National Natural Science Foundation of China (No. 72001139).
- The Guangdong Basic and Applied Basic Research Foundation (No.2023A1515011656).
- The Guangzhou-HKUST(GZ) Joint Funding Program (No.2023A03J0651).
- The Guangzhou Municipal Science and Technology Project (No.202201011235).

# Reference

Bergmann, P., K. Batzner, M. Fauser, D. Sattlegger, and C. Steger. (2021). The MVTec anomaly detection dataset: A comprehensive real-world dataset for unsupervised anomaly detection. *International Journal of Computer Vision 129* (4), 1038-1059.

Bergmann, P., M. Fauser, D. Sattlegger, and C. Steger. (2020). Uninformed students: Student-teacher anomaly detection with discriminative latent embeddings. In *Proceedings of the IEEE/CVF Conference on Computer Vision and Pattern Recognition (CVPR)*, pp. 4182-4191.

Cao, J., J. Zhang, Z. Wen, N. Wang, and X. Liu. (2017). Fabric defect inspection using prior knowledge guided least squares regression. *Multimedia Tools and Applications 76* (3), 4141-4157.

Cao, X., C. Tao, and J. Du. (2024). 3D-CSAD: Untrained 3D anomaly detection for complex manufacturing surfaces. *arXiv preprint arXiv:2404.07748*.

Çelik, A., A. Küçükmanisa, A. Sümer, A. T. Çelebi, and O. Urhan. (2022). A real-time defective pixel detection system for LCDs using deep learning-based object detectors. *Journal of Intelligent Manufacturing 33* (4), 985-994.

Hou, X. and L. Zhang. (2007). Saliency detection: A spectral residual approach. In *2007 IEEE Conference on Computer Vision and Pattern Recognition*, pp. 1-8.

Huangpeng, Q., H. Zhang, X. Zeng, and W. Huang. (2018). Automatic visual defect detection using texture prior and low-rank representation. *IEEE Access 6*, 37965-37976.

Jia, L., C. Chen, J. Liang, and Z. Hou. (2017). Fabric defect inspection based on lattice segmentation and Gabor filtering. *Neurocomputing 238*, 84-102.

Kim, M., M. Lee, M. An, and H. Lee. (2020). Effective automatic defect classification process based on CNN with stacking ensemble model for TFT-LCD panel. *Journal of Intelligent*

*Manufacturing 31* (5), 1165-1174.

Kingma, D. P. and J. Ba. (2017). Adam: A method for stochastic optimization. *arXiv preprint arXiv:1412.6980*.

Li, C., G. Gao, Z. Liu, D. Huang, and J. Xi. (2019). Defect detection for patterned fabric images based on GHOG and low-rank decomposition. *IEEE Access 7*, 83962-83973.

Liu, Y., R. T. Collins, and Y. Tsin. (2004). A computational model for periodic pattern perception based on frieze and wallpaper groups. *IEEE Transactions on Pattern Analysis and Machine Intelligence 26* (3), 354-371.

Mak, K. L. and P. Peng. (2008). An automated inspection system for textile fabrics based on Gabor filters. *Robotics and Computer-Integrated Manufacturing 24* (3), 359-369.

Mak, K. L., P. Peng, and K. F. C. Yiu. (2009). Fabric defect detection using morphological filters. *Image and Vision Computing 27* (10), 1585-1592.

Mo, D., W. K. Wong, Z. Lai, and J. Zhou. (2021). Weighted double-low-rank decomposition with application to fabric defect detection. *IEEE Transactions on Automation Science and Engineering 18* (3), 1170-1190.

Ngan, H. Y. T., G. K. H. Pang, and N. H. C. Yung. (2008). Motif-based defect detection for patterned fabric. *Pattern Recognition 41* (6), 1878-1894.

Paszke, A., S. Gross, S. Chintala, G. Chanan, E. Yang, Z. DeVito, Z. Lin, A. Desmaison, L. Antiga, and A. Lerer. (2017). Automatic differentiation in PyTorch.

Quinn, B. G. and P. J. Thomson. (1991). Estimating the frequency of a periodic function. *Biometrika 78* (1), 65-74.

Raheja, J. L., B. Ajay, and A. Chaudhary. (2013). Real time fabric defect detection system on an embedded DSP platform. *Optik 124* (21), 5280-5284.

Raheja, J. L., S. Kumar, and A. Chaudhary. (2013). Fabric defect detection based on GLCM and Gabor filter: A comparison. *Optik 124* (23), 6469-6474.

Rife, D. and R. Boorstyn. (1974). Single tone parameter estimation from discrete-time observations. *IEEE Transactions on Information Theory 20* (5), 591-598.

Shi, B., J. Liang, L. Di, C. Chen, and Z. Hou. (2019). Fabric defect detection via low-rank decomposition with gradient information. *IEEE Access 7*, 130423-130437.

Shi, B., J. Liang, L. Di, C. Chen, and Z. Hou. (2021). Fabric defect detection via low-rank decomposition with gradient information and structured graph algorithm. *Information Sciences 546*, 608-626.

Szarski, M. and S. Chauhan. (2022). An unsupervised defect detection model for a dry carbon fiber textile. *Journal of Intelligent Manufacturing 33* (7), 2075-2092.

Tao, C. and J. Du. (2023). PointSGRADE: Sparse learning with graph representation for anomaly detection by using unstructured 3D point cloud data. *IISE Transactions*, 1-14.

Tao, C., J. Du, and T.-S. Chang. (2023). Anomaly detection for fabricated artifact by using unstructured 3D point cloud data. *IISE Transactions*, 1-13.

Tsai, D. M. and C. Y. Hsieh. (1999). Automated surface inspection for directional textures. *Image and Vision Computing 18* (1), 49-62.

Yan, H., K. Paynabar, and J. Shi. (2017). Anomaly detection in images with smooth background via smooth-sparse decomposition. *Technometrics 59* (1), 102-114.

Yang, X. Z., G. K. H. Pang, and N. H. C. Yung. (2002). Discriminative fabric defect detection using adaptive wavelets. *Optical Engineering 41* (12), 3116-3126.

Yu, J., Y. Zheng, X. Wang, W. Li, Y. Wu, R. Zhao, and L. Wu. (2021). FastFlow: Unsupervised Anomaly Detection and Localization via 2D Normalizing Flows. *arXiv preprint arXiv:2111.07677*.

Zambal, S., W. Palfinger, M. Stöger, and C. Eitzinger. (2015). Accurate fibre orientation measurement for carbon fibre surfaces. *Pattern Recognition 48* (11), 3324-3332.

Zhang, H., F. Robitaille, C. U. Grosse, C. Ibarra-Castanedo, J. O. Martins, S. Sfarra, and X. P. V. Maldague. (2018). Optical excitation thermography for twill/plain weaves and stitched fabric dry carbon fibre preform inspection. *Composites Part A: Applied Science and Manufacturing 107*, 282-293.

Zhang, T. (2004). Solving large scale linear prediction problems using stochastic gradient descent algorithms. In *Proceedings of the Twenty-first International Conference on Machine Learning (ICML)*, pp. 116.

# Appendix of "A Novel Representation of Periodic Pattern and Its Application to Untrained Anomaly Detection"

## Appendix A

In order to assess the impact of penalty term $\rho_2\|\boldsymbol{\lambda_1}\|_1$ $\rho_2\|\boldsymbol{\lambda_2}\|_1$ on periodic pattern learning, we generate images with different backgrounds and input the images to the periodic pattern learning framework with different $\rho_2$.

The test images are shown in Figure 10. The background of the first image is smooth and generated using the function $\boldsymbol{Y^*}_{i,j} = \frac{0.6}{4}\left(\sin\frac{2\pi i}{40} + \cos\frac{2\pi j}{50}\right) + 0.5$. The background of the second image is non-smooth and consists of rectangles with a width of 25 and a height of 20. The values in the brighter rectangles are 0.8, and the values in the darker rectangles are 0.2. Both images have zero-mean Gaussian noise with $\sigma = 0.01$. We input these images to the periodic pattern learning farmwork with $\rho_2 = 0, \rho_2 = 100, \rho_2 = 1000$.

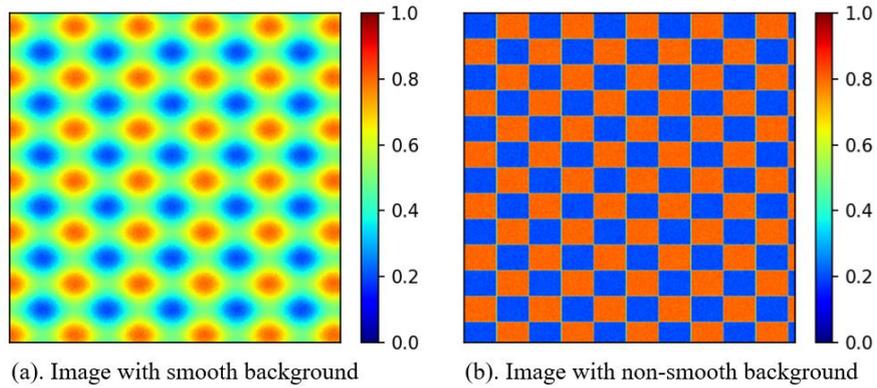

(a). Image with smooth background  (b). Image with non-smooth background

Figure 10.   Test images

The reference image construction results using PSD with the same iteration time are shown in Figure 11. The images in the first row are the reference images $\boldsymbol{Y^r}$ of the smooth image constructed by Eq. (14), using the learned $\boldsymbol{\lambda_1}$ and $\boldsymbol{\lambda_2}$ with different $\rho_2$. The images in the third row are the reference images $Y^r$ of the non-smooth image. The second row and the fourth row are the images $Y^* - Y^r$, where $Y^*$ denotes the images without any noise. The outputs of $\lambda_1$ and $\lambda_2$ are shown in Figure 12.

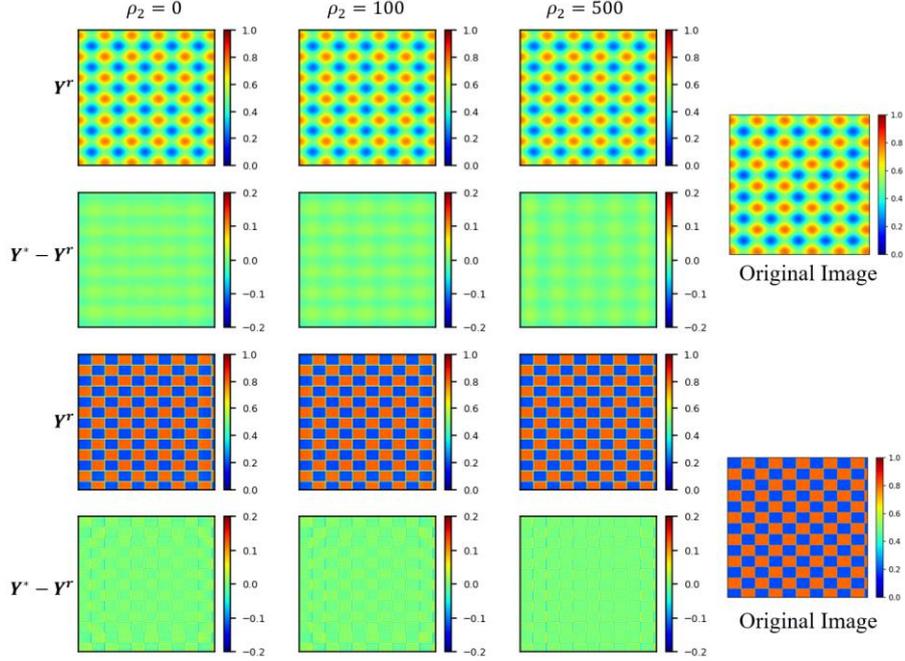

Figure 11. Reference image construction result with different $\rho_2$.

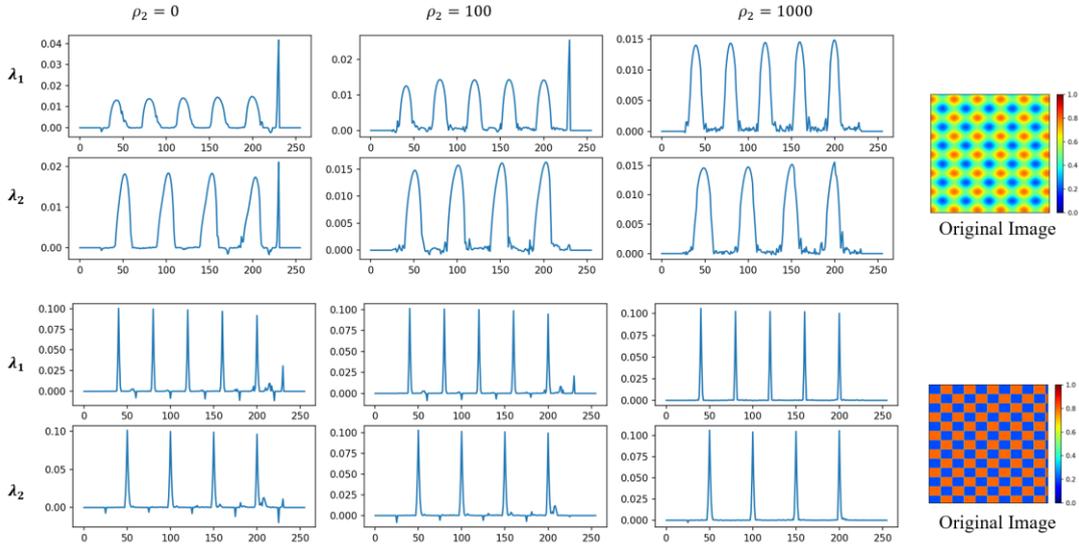

Figure 12. Results of periodic pattern $\lambda_1$ and $\lambda_2$ with different $\rho_2$

The means and standard deviations of images $Y^* - Y^r$ are listed in Table 4. Here, $\sigma^E$ denotes the standard deviation of the Gaussian noise $E$ in the original images $Y = Y^* + E$.

$\|E\|_\infty$ denotes the max absolute value in $E$. $\mu^r$ and $\sigma^r$ denote the mean and the standard deviation of the subtraction of the noise-free image $Y^*$ and reconstruction image $Y^r$ respectively. $\|Y^* - Y^r\|_\infty$ denotes the max absolute value in $Y^* - Y^r$. The results illustrate that the reconstructed images $Y^r$ are very close to the real noise-free images $Y^*$.

**Table 4.** Statistical values of differences between $Y^*$ and $Y^r$

|  |  | $\rho_2 = 0$ | $\rho_2 = 100$ | $\rho_2 = 500$ |
|---|---|---|---|---|
| Image with smooth background | $\sigma^E$ | 0.009934 | 0.009934 | 0.009934 |
|  | $\|E\|_\infty$ | 0.043500 | 0.043500 | 0.043500 |
|  | $\mu^r$ | 0.000015 | 0.000019 | 0.000011 |
|  | $\sigma^r$ | 0.010048 | 0.010866 | 0.010958 |
|  | $\|Y^* - Y^r\|_\infty$ | 0.038352 | 0.043483 | 0.047356 |
| Image with non-smooth background | $\sigma^E$ | 0.010045 | 0.010045 | 0.010045 |
|  | $\|E\|_\infty$ | 0.043792 | 0.043792 | 0.043792 |
|  | $\mu^r$ | -0.000013 | -0.000013 | -0.000012 |
|  | $\sigma^r$ | 0.029284 | 0.029379 | 0.029460 |
|  | $\|Y^* - Y^r\|_\infty$ | 0.130041 | 0.131641 | 0.129680 |

Comparing the results of the learned periodic patterns ($\lambda_1$ and $\lambda_2$) of smooth and non-smooth images, we can find that in smooth surfaces, the gradient near the optimal solution is small, making it difficult to converge to a sparse optimal solution. Comparing the results of the learned periodic patterns with different $\rho_2$, it can be concluded that using penalty $\rho_2 \|\lambda\|_1$ can help us to promote the sparsity of $\lambda$ to get better results.

## Appendix B

**Image Expansion**

In this step, the rotated image is subjected to the periodic pattern representation method, resulting in the rotated reference image $Y^*$ depicted in Figure 2.

The rotated reference image can now be expanded using the learned periodic pattern vectors. This expansion is performed row by row and column by column, inferring the values of the expanded area. Computing the pixel values of the first newly added top and bottom rows is taken as an example. $m$ denotes the length and width of the rotated reference image. $\lambda_1 \in \mathbb{R}^m$ denotes the periodic pattern vector of the column vectors of the rotated reference image. $S_1^T(\lambda_1)$ represents the first row of the weight matrix $S$ with respect to $\lambda_1$ mentioned in Section 3. $Y_B \in \mathbb{R}^{(m-1)\times m}$ denotes the first $m-1$ rows of the rotated reference image pixel matrix. The values of the newly added row $Y_0 \in \mathbb{R}^m$ at the top can be calculated as follows:

$$Y_0 = \frac{1}{\mathbf{1}_m{}^T \cdot S_1(\lambda_1)} \cdot S_1^T(\lambda_1) \cdot \begin{bmatrix} \mathbf{0}_m^T \\ Y_B \end{bmatrix}$$

$S_m^T(\lambda_1)$ represents the last row of the weight matrix $S$ with respect to $\lambda_1$. $Y_H \in \mathbb{R}^{(m-1)\times m}$ denotes the last $m-1$ rows of the image pixel matrix, and the value of the newly added row $Y_{m+1} \in \mathbb{R}^m$ at the bottom can be calculated as follows:

$$Y_{m+1} = \frac{1}{\mathbf{1}_m{}^T \cdot S_m(\lambda_1)} \cdot S_m^T(\lambda_1) \cdot \begin{bmatrix} Y_H \\ \mathbf{0}_m^T \end{bmatrix}$$